\documentclass[10pt,twocolumn,letterpaper]{article}

\usepackage{iccv}
\usepackage{times}
\usepackage{epsfig}
\usepackage{graphicx}
\usepackage{amsmath}
\usepackage{amssymb}

\usepackage{short}
\usepackage{algorithm}
\usepackage{algorithmic}
\usepackage{booktabs}
\usepackage{subfig}
\usepackage{multirow}
\usepackage{makecell}

\newcommand{\method}{ACKD\xspace}

\usepackage[pagebackref=true,breaklinks=true,letterpaper=true,colorlinks,bookmarks=false]{hyperref}

\iccvfinalcopy % *** Uncomment this line for the final submission

\def\iccvPaperID{3070} % *** Enter the ICCV Paper ID here

\ificcvfinal\fi

\begin{document}

%%%%%%%%% TITLE
\title{Improved Knowledge Distillation via Adversarial Collaboration}
% \title{$A^2CKD$: Adversarial Attentive Collaborative Knowledge Distillation}

\author{Zhiqiang Liu$^{1}$\thanks{Co-first authorship.}\quad Chengkai Huang$^{2*}$\quad Yanxia Liu$^1$\thanks{ Corresponding author.}\\
$^1$South China University of Technology\\
$^2$The University of New South Wales\\
{\tt\small sezhiqiangliu@mail.scut.edu.cn, chengkai.huang1@unsw.edu.au, cslyx@scut.edu.cn}
}

\maketitle
\ificcvfinal\thispagestyle{empty}\fi

%%%%%%%%% ABSTRACT
\begin{abstract}
   Knowledge distillation has become an important approach to obtain a compact yet effective model. To achieve this goal, a small student model is trained to exploit the knowledge of a large well-trained teacher model. However, due to the capacity gap between the teacher and the student, the student's performance is hard to reach the level of the teacher. Regarding this issue, existing methods propose to reduce the difficulty of the teacher's knowledge via a proxy way. We argue that these proxy-based methods overlook the knowledge loss of the teacher, which may cause the student to encounter capacity bottlenecks. In this paper, we alleviate the capacity gap problem from a new perspective with the purpose of averting knowledge loss. Instead of sacrificing part of the teacher's knowledge, we propose to build a more powerful student via adversarial collaborative learning. To this end, we further propose an Adversarial Collaborative Knowledge Distillation (ACKD) method that effectively improves the performance of knowledge distillation. Specifically, we construct the student model with multiple auxiliary learners. Meanwhile, we devise an adversarial collaborative module (ACM) that introduces attention mechanism and adversarial learning to enhance the capacity of the student. Extensive experiments on four classification tasks show the superiority of the proposed ACKD.
\end{abstract}

%%%%%%%%% BODY TEXT
\section{Introduction}
Convolutional neural networks (CNNs) have achieved impressive success in many vision tasks such as image classification~\cite{he2016deep, huang2017densely, sandler2018mobilenetv2}, object detection~\cite{lin2017focal, ren2015faster, redmon2016you}, and image generation~\cite{Goodfellow2014GenerativeAN, Liu2021CTSF}. To pursue high performance, the design of CNNs tends to be more and more complicated. These structures demand huge computation and storage resources, which are unavailable in resource-limited devices like mobile phones. In order to address this problem, many researchers propose a number of model compression techniques such as lightweight model design~\cite{sandler2018mobilenetv2, ma2018shufflenet, hu2018squeeze}, model quantization~\cite{wu2016quantized, jacob2018quantization, Xie2020DeepTQ}, and model pruning~\cite{he2017channel, zhuang2018discrimination, chin2020towards}. In recent years, knowledge distillation~\cite{hinton2015distilling, romero2014fitnets, yuan2020revisiting} has attracted widespread attention, which aims to improve the performance of a small model (student) by exploiting the knowledge of a large well-trained model (teacher).

Existing knowledge distillation methods mainly focus on what kind of knowledge the teacher has learned (\eg, logit-based~\cite{hinton2015distilling}, feature-based~\cite{romero2014fitnets} and relation-based~\cite{peng2019correlation} knowledge) and how to transfer the knowledge from the teacher to the student (\ie, proper loss function). However, the teacher-student pairs are sometimes significantly different (in structure and size), which leads to extremely learning difficulty for the student. To alleviate this problem, some works~\cite{cho2019efficacy, mirzadeh2019improved} seek the proxy teachers to reduce the difficulty of the teacher's knowledge. Specifically, Cho \etal~\cite{cho2019efficacy} propose that the student can gain better performance through learning the knowledge from an early-stopped teacher model rather than final converged one. Recently, Mirzadeh \etal~\cite{mirzadeh2019improved} first transfer the knowledge to medium capacity assistant models, then distill knowledge from assistants to the student, which is multi-step knowledge distillation framework. 

However, although these proxy-based methods ease optimization difficulty for the student to a certain extent, there still remain the following problems. First, the core knowledge of teacher might be lost, which would prevent the student from learning useful information in the distillation process. The key idea of the early-stopped and assistant teachers is to reduce the teacher's capacity such that the teacher's knowledge can be accepted by the student. In this process, they lose part of knowledge from the original teacher, which will be discussed in detail in section~\ref{sec:analysis}. Second, the whole training process is complicated and time-consuming. These methods require lots of grid searches and additional training to obtain an optimal early-stopped anchor or teacher assistant in different knowledge distillation scenarios.

To avoid the above issues, in this paper, we provide a new perspective to alleviate the aforementioned learning difficulty problem. Rather than rudely reduce the difficulty of the teacher's knowledge, we advocate strengthening the student's capacity. To achieve this, we exploit the idea of collaborative learning that many collaborators/learners form a learning group in order to reach the same goal. Given multiple small-sized students, we can effectively exploit the knowledge of the teacher via ensemble learning. However, such cumbersome student models violate the purpose of model compression. Thus, how to form a learning group while avoiding incurring additional parameter overhead in inference remains a question.

In this paper, for the sake of improving the student's capacity while keeping the student be efficiency in inference, we propose an Adversarial Collaborative Knowledge Distillation (\method) method that can significantly improve the performance of knowledge distillation. In the training phase, we build the student model with multiple auxiliary learners. These learners form a learning group and collaborate to achieve the same goal that fully exploits the knowledge of the teacher. For the purpose of achieving better collaboration, we introduce the attention mechanism to filter the information of auxiliary learners. Moreover, to enhance the ability of auxiliary learners to learn more comprehensive knowledge, the auxiliary learners are urged to be diverse. To this end, we distinguish their distributions using adversarial learning via discriminators. As a result, the proposed student model is more powerful to exploit the teacher's knowledge. In the inference phase, we discard all auxiliary learners such that the final student is compact and effective. Extensive experiments on four benchmarks demonstrate the effectiveness of proposed \method.

Our contributions are summarized as follows:
\begin{itemize}
    \item We propose an Adversarial Collaborative Knowledge Distillation (\method) method that effectively alleviates the capacity gap problem in knowledge distillation. Unlike existing methods that rudely reduce the difficulty of the teacher's knowledge, we alleviate this problem from a new perspective of strengthening the student's capacity.
    \item We propose an adversarial collaborative learning strategy that effectively improves the student's capacity. To this end, we introduce the attention mechanism to recognize the importance of different auxiliary learners. Moreover, we introduce the adversarial learning to promote representational diversity.
    \item To investigate the effectiveness of our proposed method, we conduct extensive experiments on four classification datasets. These results demonstrate that our \method outperforms state-of-the-art methods.
\end{itemize}

\section{Related Work}
\textbf{Knowledge Distillation.} Knowledge distillation is mainly applied to the field of model compression~\cite{han2015deep, rastegari2016xnor, ye2019student, liu2021semi}, which trains a small and effective student model with the help of large teacher models' knowledge. As a pioneer work, Hinton \etal~\cite{hinton2015distilling} take output probabilities as knowledge, and the student learns these knowledge by minimizing Kullback–Leibler (KL) divergence. After that, Romero \etal~\cite{romero2014fitnets} exploit the intermediate representations' knowledge by matching the feature activations of the teacher and the student. Besides, Zagoruyko \etal~\cite{Zagoruyko2017PayingMA} introduce attention mechanism to the knowledge transfer, and improve the student's performance by mimicking attention feature map. Recently, some works define the knowledge according to the relation between layers or samples. Specifically, Tung \etal~\cite{tung2019similarity} urge the student to learn how to preserve the pairwise similarities. Tian \etal~\cite{tian2019contrastive} introduce contrastive objectives by constructing positive and negative pairs. However, most existing works ignore the capacity gap between the teacher and the student, which may result in performance degradation~\cite{cho2019efficacy, jin2019knowledge}. To address this problem, Cho \etal~\cite{cho2019efficacy} reduce the gap via an early-stopped teacher, and Mirzadeh \etal~\cite{mirzadeh2019improved} introduce teacher assistants to simplify the teacher's knowledge. Unlike these methods, in this paper we bridge the gap from the perspective of improving the student's capacity.

\textbf{Collaborative Learning.} Collaborative learning refers to two or more learners work together to solve a common task~\cite{dillenbourg1999collaborative}. For that purpose, learners learn from based on their diverse strengths such that they reach goals more efficiently. Recently, some studies have introduced this idea to improve the training efficiency of deep neural networks with knowledge distillation~\cite{Lan2018KnowledgeDB, li2020online}. To be specific, Song \etal~\cite{song2018collaborative} design a hierarchical structure of multiple branches, which are trained simultaneously to improve backbone's performance. Chen \etal~\cite{chen2020online} introduce self-attention to improve the branch diversity and distill the knowledge of auxiliary peers into main branch. Additionally, some works apply collaborative learning across networks~\cite{chen2020online, chung2020feature}. Zhang \etal~\cite{Zhang2018DeepML} propose a deep mutual learning training fashion, in which all networks distill knowledge to each other simultaneously. Guo \etal~\cite{DBLP:conf/cvpr/GuoWWYLHL20} train networks under the supervision of ensemble targets. In this paper, we follow the idea of collaborative learning to improve the imitation ability of the student network in knowledge distillation.

\textbf{Deep Supervision.} Deep supervision is proposed to alleviate the gradient vanish problem in deep neural networks by attaching auxiliary classifier to intermediate layers~\cite{lee2015deeply, szegedy2015going}. These auxiliary classifiers will be removed in the testing phase to keep the efficiency of the model. This training mechanism has been widely proved to be effective in different applications, such as edge detection~\cite{xie2015holistically}, object detection~\cite{liu2016ssd}, and semantic segmentation~\cite{zhao2017pyramid}. Recently, some works~\cite{yao2020knowledge, zhang2020task} apply this idea to the field of knowledge distillation. In this paper, we extend this idea to construct multiple auxiliary learner to collaboratively learning knowledge from the teacher network.

\textbf{Adversarial Learning.} Generative adversarial networks~\cite{goodfellow2014generative} is a generative model framework, which consists of a generator and a discriminator. The generator generates fake images from the noise to deceive the discriminator while the discriminator tries to distinguish the real and fake images. Recently, some works~\cite{shen2019meal, chung2020feature, zhang2020amln} introduce this adversarial learning strategy to discriminate the features generated by different networks. In this paper, we regard the auxiliary learners as generators and use discriminators to distinguish the output features between different learners. In this way, we improve the expressive diversity of different auxiliary learners.

\section{Proposed Method}

\begin{table}[t]
\begin{center}
\resizebox{0.48\textwidth}{!}{
\begin{tabular}{c|cc}
        \toprule 
        Method & ESKD~\cite{cho2019efficacy} & TAKD~\cite{mirzadeh2019improved}\\
        \midrule
        Teacher (T) & WideResNet-28-4 & WideResNet-28-4\\
        Proxy (P) & WideResNet-28-4 (65)  & WideResNet-28-2 \\
        Student (S) & resnet56 & resnet56\\
        \midrule
        Acc(T) & 78.91 & 78.91 \\
        Acc(P) & 76.30 & 76.78 \\
        AccGap(T, P) & 2.61 & 2.13 \\
        \midrule
        CKA(T, P)  & 0.7761 & 0.8234\\
        CKA(P, S)  & 0.8016 & 0.8113\\
        CKA(T, S)  & 0.7695 & 0.7770\\
        \midrule
        KL(T, P)  & 0.6627 & 0.7710\\
        KL(P, S)  & 0.8495 & 0.8391\\
        KL(T, S)  & 0.9319 & 0.9166\\
		\bottomrule
		\end{tabular}}
\end{center}
	\caption{Comparasions of different proxy methods. Acc($\cdot$) denotes Top-1 accuracy ($\%$). AccGap(T, P) = Acc(T) - Acc(P). CKA($\cdot$,$\cdot$) and KL($\cdot$,$\cdot$) denote CKA similarity~\cite{kornblith2019similarity} and Kullback–Leibler divergence between two networks, respectively. The number inside the parentheses is the total number of training epochs with early-stopped strategy.}
		\label{tab:proxy}
\end{table}

In this section, we will introduce our proposed method in detail. In section~\ref{sec:analysis}, we analyze the advantages and shortcomings of existing proxy-based ways to motivate our method. In section~\ref{sec:acl}, we introduce the overview of proposed method that improves the performance of the student via an adversarial collaboration learning strategy. Then, we introduce the optimization and the whole knowledge flow paradigm in our framework in sections~\ref{sec:t-a}, \ref{sec:ac}, and \ref{sec:self-distillation}.
\subsection{Analysis on Proxy-based Approaches}\label{sec:analysis}
The vanilla KD method is a two-stage process in which a high capacity teacher model is trained and then used for distillation. However, due to the high capacity divergence between the teacher and the student, it results in a common phenomenon that the student is hard to reach the teacher's performance after the distillation process.

To alleviate this divergence, many researchers try to distill knowledge from the teacher to the student in a more smooth way. One of the most well-known methods is TAKD~\cite{mirzadeh2019improved}, which utilizes proxy models to alleviate the gap, aiming to provide an easier learning curve for the student. Another insightful method is to reduce the teacher's capacity by using early-stop teacher regularization~\cite{cho2019efficacy}, in this way, the student can gain better performance. 

A major advantage of these methods is to smooth the student's learning curve and mitigate the mismatched capacity between teacher and student. However, there are certain drawbacks associated with the use of the proxy model and early-stop teacher model. Firstly, although these approaches alleviate the student's learning burdens, they also lose abundant knowledge of the original teacher in the distillation process, which may cause that the student model can not 
draw some key information. As Table~\ref{tab:proxy} showed, we use qualitative analysis in order to gain insights into the information loss in the distillation process. We conduct experiments with the teacher-student pair of WRN-28-4 and resnet56 on CIFAR-100. Specifically, we use both CKA~\cite{kornblith2019similarity} and KL divergence to measure the representational similarity among the teacher, the proxy, and the student. The larger the CKA, the more similar the representations among the teacher, the proxy, and the student. The larger the KL divergence, the greater the difference between the two distributions. Besides, we use the relative accuracy (AccGap) to evaluate the capacity gap among the teacher and the proxy. From Table~\ref{tab:proxy}, the CKA of the teacher-proxy pair is higher than the teacher-student pair while KL is quite the reverse, which means that the student model suffers the loss of information during the proxy distillation process. Meanwhile, both AccGap of ESKD and TAKD are more than 2\%, also proving the existence of capacity gap and information loss between teacher and proxy model. Secondly, these methods require lots of grid searches to obtain an optimal proxy model (\ie, early-stopped anchor or teacher assistant), which is time-consuming and inefficient.

Thus, how to maintain the teacher’s capacity (original abundant information) while improving the students’ capacity has become a prisoner’s dilemma problem: reducing the difficulty of student learning would hurt teachers' capacity while a large and well-trained teacher would mismatch its student capacity.  

\begin{figure*}[t]
    \centering
	\includegraphics[width=0.9\linewidth]{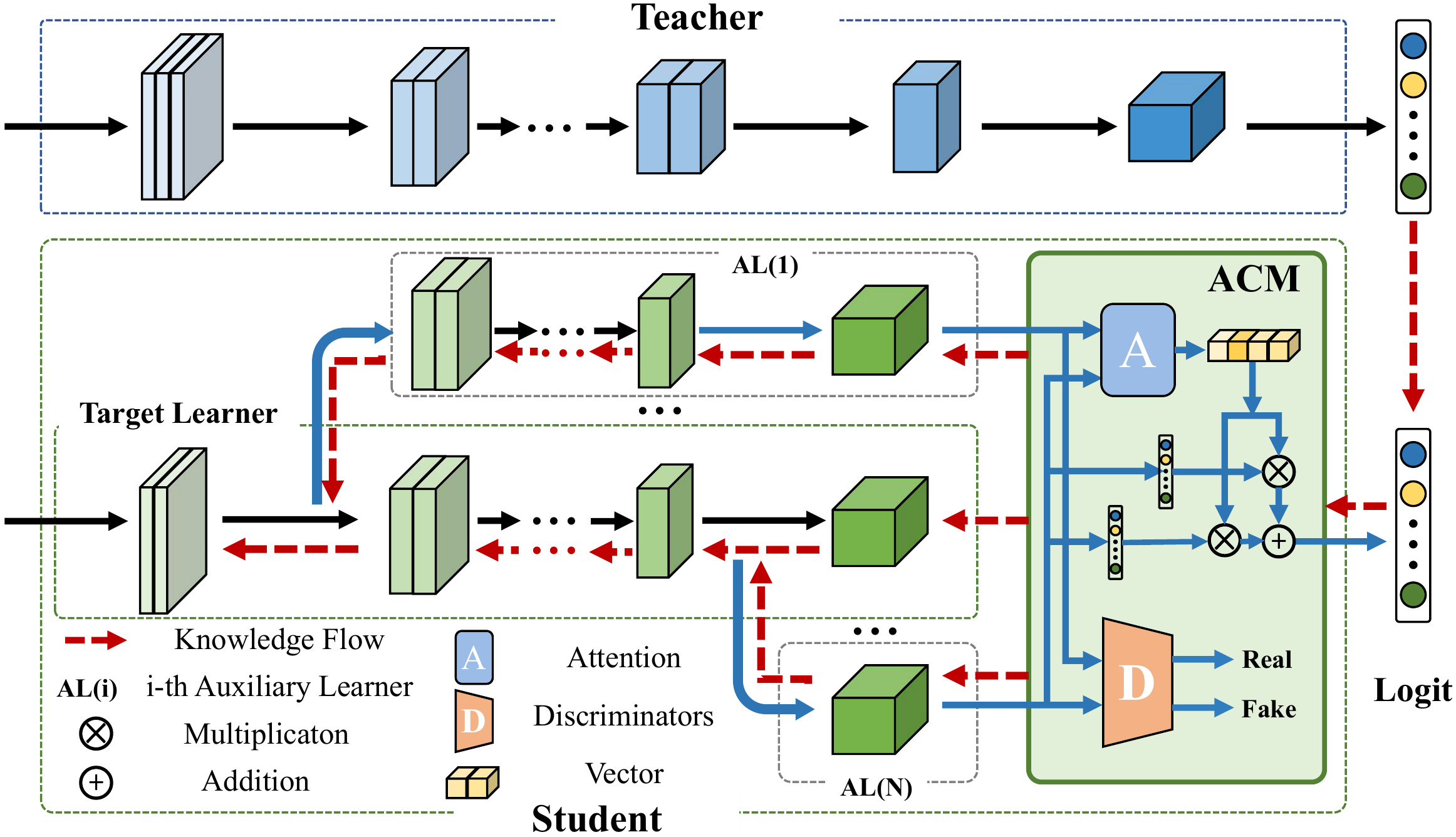}
	\caption{An overview of the proposed \method. In the training phase, we first construct the student model with auxiliary learners (gray boxes). Then, we improve the student's capacity via an adversarial collaborative learning strategy. In the inference phase, we only use the target learner and remove all auxiliary learners. (Best viewed in color)}
	\label{fig:framework}
\end{figure*}

\subsection{Adversarial Collaborative Learning}\label{sec:acl}
Imagine a real-world scenario where incomprehensible (complex) knowledge is given by the teacher model, and the student is asked to absorb these contents by its own small-capacity network. The capacity gap discussed in the previous subsection occurs and makes the knowledge distillation less effective than it could be. To alleviate this issue, we seek to exploit the idea of collaborative learning that different learners work together to solve a common task. Inspired by this, we propose to strengthen the student’s capacity by forming a learning group rather than rudely reduce the difficulty of the teacher's knowledge. In this way, it is possible to achieve better distillation performance without damaging the teacher's original capacity and semantic information.

To form a learning group that exploits the knowledge of the teacher, we can build a multi-students~\cite{you2018learning} ensemble learning framework~\cite{dietterich2000ensemble}. However, such a cumbersome model is inconsistent with our goal of model compression. In order to address this problem, in this paper, we propose to construct the student model with multiply branches. For convenience, we call these branches as auxiliary learners, and use them to form a learning group. The backbone model without auxiliary branches is named as target learner. In practice, we select different layers branch from the target learner as our auxiliary learners' inputs. The auxiliary learners are composed of different high-level feature extractor blocks. In this way, the auxiliary learners share the low-level feature information but have their own high-level semantic information. 

The auxiliary learners are asked to reach the same objective that exploits the knowledge of the teacher in the greatest degree. To fully extract the representation information of the auxiliary learners, we introduce an attention mechanism to collaborative learning. In this way, each auxiliary is urged to focus on what it does well. However, auxiliary learners share the same low-level layers, which might lead to a lack of diversity in expressivity. To alleviate this issue, we introduce adversarial learning to promote the diversification of auxiliary learners. In the whole training phase, we hope that the auxiliary learners are in the state of \textit{collaborative learning} while exploring unique representation via \textit{adversarial learning}. In the inference phase, we discard all auxiliary learners to obtain a compact and effective model. For simplicity, we call our training framework as Adversarial Collaborative Knowledge Distillation (\method).

Figure~\ref{fig:framework} shows the overall architecture. In practice, the teacher's knowledge first flows to auxiliary learners. Then, the auxiliary learners mine this knowledge via an adversarial collaborative manner. In this process, the knowledge flows to shared low-level layers of the target learner. Besides, the target learner exploits the knowledge from auxiliary learners, which is a more smooth learning manner than directly learning from the teacher. In the following, we will describe in detail how the knowledge flows and the parameters update in our framework.

\subsection{Teacher-Auxiliary Learners Knowledge Flow}\label{sec:t-a}
We follow the common KD approach to distill teacher's knowledge into auxiliary learners. Suppose that we train the models on the dataset $\mD = \{(x_i, y_i)\}_{i=1}^{N}$, where $N$ is the number of samples. We can obtain the logits $z_i$ of $i^{th}$ sample. Then we can get the softened output probability of a model by applying a temperature factor as follows:
\begin{equation}\label{eq:soft_target}
    p_i=\frac{exp(z_i^j/\tau)}{\sum_{j=1}^{C}exp(z_i^j/\tau)}, 
\end{equation}
where $\tau$ is a temperature factor. In this way, we can get $p_i^t$ of the teacher and $p_i^a$ of auxiliary learners. The auxiliary learners learn the knowledge of the teacher by minimizing Kullback-Leibler (KL) divergence between these two output distributions as follows:
\begin{equation}\label{eq:kl}
    \mL_{kd}^a=KL(p^a,p^t)=\sum_{x\sim \mD}p^a\log\frac{p^a}{p^t}.
\end{equation}

Taking advantage of collaborative learning, the output distribution of auxiliary learners is easier to fit that of the teacher. Through backward propagation, we are able to get more robust low-level representations of the target learner, which is conducive to the target learner's optimization.

\subsection{Adversarial Collaborative Module}\label{sec:ac}
In this subsection, we introduce how the auxiliary learners collaborate to exploit the teacher's knowledge via an Adversarial Collaborative Module (ACM), as shown in Figure~\ref{fig:acm}. For simplicity, we use two auxiliary learners as examples for illustration. In ACM, there are two complementary learning steps: attention-based collaborative learning and adversarial-based diversity learning.

\subsubsection{Attention-Based Collaborative Learning}
To take advantage of different auxiliary learners' ability, we present the attention-based collaborative module to enhance the representations of auxiliary learners. First, we concatenate all feature maps from different auxiliary learners. Then, we input the fused feature into a two-layer perceptron with a softmax to recognize the importance of different learners. Formally, the attention weight vector and output logits of the learning group can be computed as follows:
\begin{equation}\label{eq:attention}
\begin{split}
    \bx &= g({\rm Concat}(h^{(1)}, .., h^{(M)})), \\
    \ba^{(k)} &= {\rm Softmax}({\rm MLP}(\bW_m, \bx)), \\
    z &= \sum_{k=1}^M\ba^{(k)}z^{(k)}, 
\end{split}
\end{equation}
where $M$ is total number of auxiliary learners; $h^{(k)}$ denotes the $k^{th}$ auxiliary learner's feature map; $g(\cdot)$: $\mathbb{R}^{C \times H \times W} \rightarrow \mathbb{R}^{C \times HW}$ flattens the feature maps in spatial dimension; ${\rm \bW}_{m}$ denotes the parameters of ${\rm MLP}$; $\ba^{(k)}$ is the attention weight of $k^{th}$ auxiliary learner. After we get attention-based logit, we compute output probability in Eqn~\ref{eq:soft_target}. We update the parameters of auxiliary learners and ${\rm MLP}$ in Eqn~\ref{eq:kl}.

\subsubsection{Adversarial-Based Diversity Learning}

\begin{figure}[t]
    \centering
	\includegraphics[width=1\linewidth]{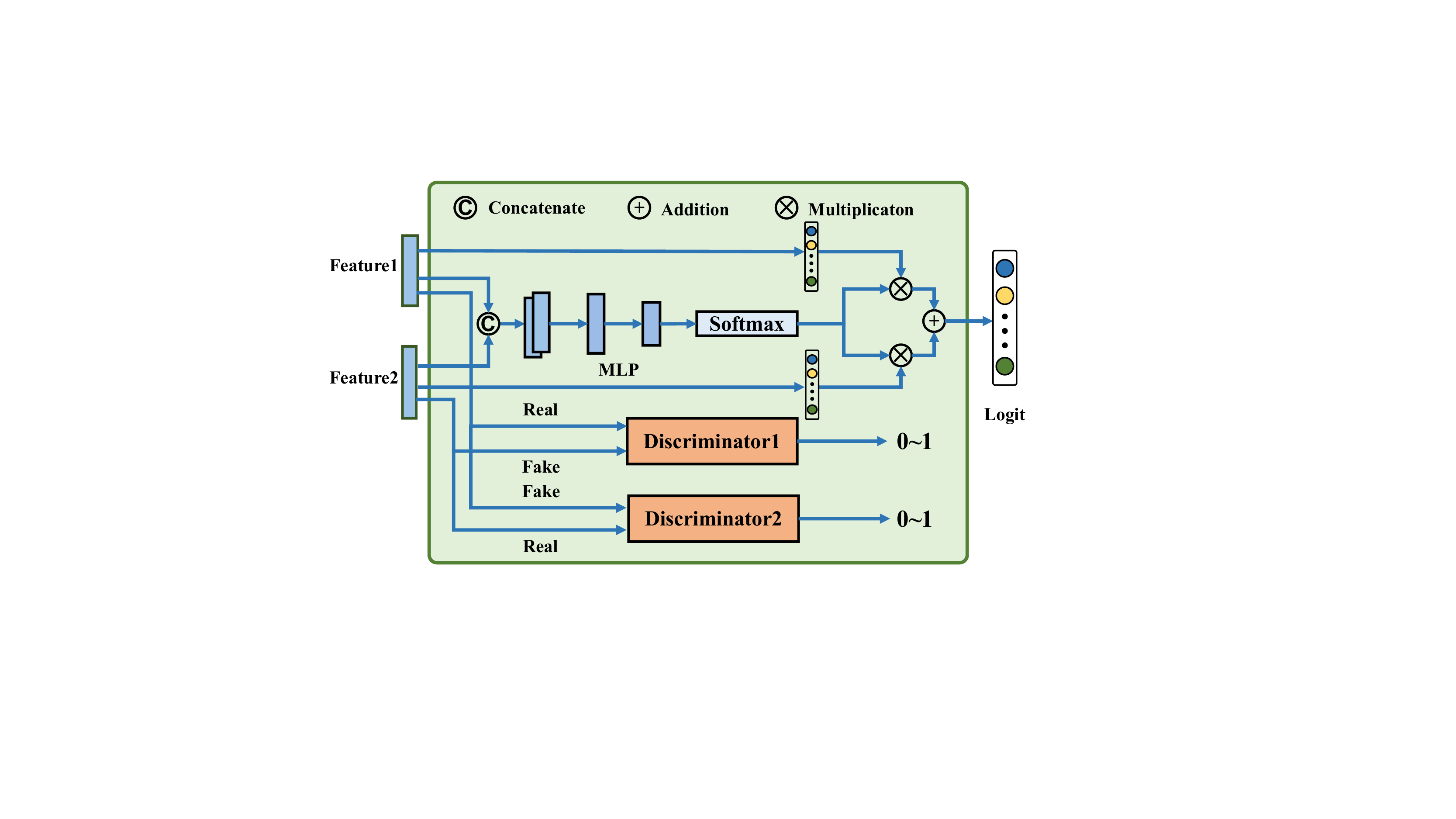}
	\caption{An overview of the proposed ACM. }
	\label{fig:acm}
\end{figure}

Although each auxiliary learner has its own branch structure, their feature inputs come from the target learner and the knowledge learned in the auxiliary learners is easy to overlap between each other. It is still difficult to avoid the homogeneity of their network parameters, which may hurt the performance of the ensemble result and even distillation performance. To alleviate this problem, we leverage the adversarial training to force auxiliary learners to learn diverse feature distribution. As Figure~\ref{fig:acm} illustrated, each learner has a corresponding discriminator aiming at distinguishing whether the feature output belongs to their own. 

Following the traditional adversarial training definition, we regard the auxiliary learners' network feature as generative distribution and name the $i^{th}$ learner's discriminator as $D_i$. The architecture of the discriminator is a simple stack of $1\times1$ Conv with ReLU. Each auxiliary learner tries to fool the discriminator while the discriminator need to discriminate the right source of feature. Our overall adversarial loss for $i^{th}$ discriminator is as follows:
\begin{equation}\label{eq:adi}
    \mL_{ad}^i=\frac{1}{M-1}\sum_{j\neq i}\mathop{\mathbb{E}}\limits_{x\sim p(i)}{\rm log}D_i(x)+\mathop{\mathbb{E}}\limits_{x\sim p(j)}{\rm log}(1-D_i(x)),
\end{equation}
where $p(i)$ denotes the output feature produced by $i^{th}$ auxiliary learner, $M$ is total number of auxiliary learners. Final adversarial loss is a average of the individual ones:
\begin{equation}\label{eq:ad}
    \mL_{ad}=\frac{1}{M}\sum_{i=1}^M\mL_{ad}^i.
\end{equation}

Through adversarial training manners, we can make better use of the diversity of semantic information contained in the auxiliary learners. Meanwhile, the parameters of target learner also are updated in this process.

\begin{table*}[t]
\begin{center}
\begin{tabular}{c|ccccc}
			\toprule 
			Teacher & WideResNet-40-2 & resnet110 & MobileNetV2 (1.4) & ShuffleNetV1 & WideResNet-28-4  \\
			Student & WideResNet-40-1 & resnet20 & MobileNetV2 (0.5) & MobileNetV2 (0.5) & resnet56 \\
			\midrule
			Teacher & 76.22 & 74.00 & 69.57 & 71.85 & 78.91 \\
			Student & 71.78 & 69.20 & 65.17 & 65.17 & 73.05 \\
		    \midrule
			AT~\cite{Zagoruyko2017PayingMA} & 72.94 & 70.78 & 66.26 & 65.55 & 73.64 \\
			SP~\cite{tung2019similarity} & 73.18 & 70.69 & 61.52 & 67.67 & 74.28 \\
			CC~\cite{peng2019correlation} & 72.22 & 70.30 & 65.39 & 65.16 & 73.16 \\
			VID~\cite{ahn2019variational} & 72.52 & 70.26 & 65.15 & 65.49 & 73.24 \\
			CRD~\cite{tian2019contrastive} & 73.57 & 70.88 & 66.47 & 67.32 & 74.34\\
           	SemCKD~\cite{chen2020cross}& 73.32 & 70.41 & 67.70 & 67.19 & 73.13\\
			KD~\cite{hinton2015distilling} & 73.54 & 70.73 & 67.94 & 68.19 & 74.21 \\
			ESKD~\cite{cho2019efficacy} & 73.21 & 70.69 & 67.61 & 67.82 & 74.61\\
			TAKD~\cite{mirzadeh2019improved} & 73.52 & 71.42 & 68.88 & 68.71 & 75.29 \\
			\midrule
		    \method (Ours) & \textbf{75.29} & \textbf{72.38} & \textbf{69.43} & \textbf{69.48} & \textbf{75.91}\\
		
		    \bottomrule
		\end{tabular}
\end{center}
	\caption{Comparison with state-of-the-art models on CIFAR-100.}
		\label{tab:cifar-100}
\end{table*}

\begin{table}[t]
\begin{center}
\begin{tabular}{c|cc}
			\toprule 
			Teacher & resnet110 &  WideResNet-28-4  \\
			Student & resnet20 & resnet56 \\
			\midrule
			Teacher & 84.62 & 87.74\\
			Student & 82.29 & 83.95\\
		    \midrule
			AT~\cite{Zagoruyko2017PayingMA} & 83.84 & 85.16 \\
			SP~\cite{tung2019similarity} & 79.75 & 84.34 \\
			CC~\cite{peng2019correlation} & 83.89 & 85.14 \\
			CRD~\cite{tian2019contrastive} & 84.11 & 84.99 \\
% 			SSKD~\cite{xu2020knowledge} & 83.88 & 85.31 \\
            SemCKD~\cite{chen2020cross} & 84.13&85.12\\
			KD~\cite{hinton2015distilling} & 84.03 & 85.18\\
			ESKD~\cite{cho2019efficacy} & 83.78 & 85.44 \\
			TAKD~\cite{mirzadeh2019improved} & 84.21 & 85.37\\
			\midrule
		    \method (Ours) & \textbf{84.42} & \textbf{85.63}\\
		
		    \bottomrule
		\end{tabular}
\end{center}
	\caption{Comparison with state-of-the-art models on CINIC-10.}
		\label{tab:cinic-10}
\end{table}

\subsection{Auxiliaries-Target Learner Knowledge Flow}\label{sec:self-distillation}

After the adversarial collaborative training, the auxiliary learners gain diverse and rich semantic information. Thus, it benefits the optimization of the target learner via shared low-level layers. Meanwhile, we dig these knowledge via a self-distillation manner (\ie, minimizing the distribution divergence between the target learner and auxiliary learners). We formulate it as follows:
\begin{equation}\label{eq:kl_S}
    \mL_{kd}^s=\sum_{x\sim \mD}p^s\log\frac{p^a}{p^s},
\end{equation}
where $p^s$ and $p^a$ are the output probability of the target learner and auxiliary learners, respectively. Such optimization is much easier than the vanilla KD process. The main reason is that the auxiliary learners have same structure with the target learner, and the target learner shares low-level layers with auxiliary learners. This special characteristic of structure makes the target learner be easier to understand the knowledge of auxiliary learners.

The overall training objective of \method is as follows:
\begin{equation}\label{eq:total}
    \mL=\lambda_1\mL_{kd}^a + \lambda_2\mL_{kd}^s + \lambda_3\mL_{ad} + \lambda_4\mL_{ce},
\end{equation}
where $\lambda_1$, $\lambda_2$, $\lambda_3$, $\lambda_4$ are balanced hyperparameters, we set them to 1 in our experiments. $\mL_{ce}$ denotes typical cross entropy loss for the target learner and auxiliary learners.

\section{Experiments}
\subsection{Datasets and Networks}
\textbf{Dataset.} We conduct experiments on four widely used benchmarks. 1) \textit{CIFAR-100} is composed of 50K training images and 10K testing images, which are divided into 100 fine-grained categories. 2) \textit{CINIC-10} consists of 90K training and 90K testing images, which are from CIFAR and ImageNet datasets. This dataset is more complex than CIFAR dataset. 3) \textit{Tiny-ImageNet} has 200 classes, each of which contains 500 images for training and 50 images for validation. All images are at the resolution of $64\times64$. 4) \textit{ImageNet} is a large-scale dataset with 1000 classes, which contains 1.2 million images for training and 50k images for validation. All images are cropped at the resolution of $224\times224$ for training and validation.

\begin{table*}[h]
\begin{center}
\resizebox{0.98\textwidth}{!}{
\begin{tabular}{c|cc|cccccccc|c}
\toprule 
        & teacher & student &
          AT~\cite{Zagoruyko2017PayingMA} &  CC~\cite{peng2019correlation} &SP~\cite{tung2019similarity} &
          CRD~\cite{tian2019contrastive} &
        %   SSKD~\cite{xu2020knowledge} &
        SemCKD~\cite{chen2020cross}&
          KD~\cite{hinton2015distilling} & 
          ESKD~\cite{cho2019efficacy} & 
          TAKD~\cite{mirzadeh2019improved} &
          \method (Ours)\\
			\midrule
            Top-1 & 52.98 & 51.60 & 51.59 & 51.82 & 53.71 & 54.33 & 54.19 & 54.20 & 52.58 & 54.95 & \textbf{55.97} \\
			Top-5 & 75.97 & 74.37 & 76.96 & 75.13 & 76.83 & 77.52 & 77.48 & 77.45 & 76.77 & 77.44 & \textbf{78.15} \\
			\bottomrule
		\end{tabular}}
\end{center}
	\caption{Test Accuracy ($\%$) of different methods on Tiny-ImageNet.}
		\label{tab:tiny-imagenet}
\end{table*}

\begin{table*}[h]
\begin{center}
\resizebox{0.98\textwidth}{!}{
\begin{tabular}{c|cc|cccccccc|c}
\toprule 
        & teacher & student &
          AT~\cite{Zagoruyko2017PayingMA} & SP~\cite{tung2019similarity} & CC~\cite{peng2019correlation} & ONE~\cite{Lan2018KnowledgeDB} &
          CRD~\cite{tian2019contrastive} & SemCKD$^\dagger$~\cite{chen2020cross} & KD~\cite{hinton2015distilling}&ESKD$^\dagger$~\cite{cho2019efficacy} & 
          \method (Ours)\\
			\midrule
			Top-1 & 26.69 & 30.25 & 29.30 & 29.38 & 30.04 & 29.45 & 28.83 & 29.13 & 29.34 &29.12 &\textbf{28.67} \\
			Top-5 & 8.58 & 10.93 & 10.00 & 10.20 & 10.83 & 10.41 & 9.87 & -&10.12&- & \textbf{9.57} \\
			\bottomrule
		\end{tabular}}
\end{center}
	\caption{Test error ($\%$) of different methods on ImageNet. $^\dagger$The authors only report Top-1 error.}
		\label{tab:ImageNet}
\end{table*}

\textbf{Network Constructions.} We perform experiments with four different types of networks, including WideResNet~\cite{zagoruyko2016wide}, resnet~\cite{he2016deep}, MobileNetV2~\cite{sandler2018mobilenetv2}, and ShuffleNetV1~\cite{zhang2018shufflenet}. Specifically, WideResNet-d-w represents the WideResNet with depth $d$ and width factor $w$. MobileNetV2 (w) denotes MobileNetV2 with a width multiplier of $w$. To investigate the generalization ability of our method, we construct teacher-student pairs with similar structures (\eg, resnet110-resnet20) and different structures (\eg, ShuffleV1-MobileNetV2).

\textbf{Compared Methods.} We compare our proposed \method with many state-of-the-art methods, including KD~\cite{hinton2015distilling}, 	AT~\cite{Zagoruyko2017PayingMA}, SP~\cite{tung2019similarity},
CC~\cite{peng2019correlation},
VID~\cite{ahn2019variational}, CRD~\cite{tian2019contrastive}, and
SemCKD~\cite{chen2020cross}. Besides, we compare our method with ESKD~\cite{cho2019efficacy} and TAKD~\cite{mirzadeh2019improved}, which also aims to alleviate the capacity gap. The implementations of compared methods are mainly based on author-provided codes and a open-resource benchmark~\cite{tian2019contrastive}. As the authors' suggestions, we set the number of early-stopped training epochs to 65 (on CIFAR-100 and CINIC-10) and 35 (on Tiny-ImageNet and ImageNet) for ESKD. We choose the model, whose performance is close to average performance of the teacher and the student, as teacher assistant for TAKD (\eg, we regard resnet44 (71.98\%) as assistant model for the pair of resnet110 (74.00\%) and resnet20 (69.20\%) on CIFAR-100).

\textbf{Implementation Details.} On CIFAR-100 and CINIC-10 datasets, we run a total of 200 epochs for all methods with SGD optimizer. We set batch size to 128, momentum to 0.9, and weight decay to 5e-4. We initialize the learning rate to 0.1 (MobileNetV2 and ShuffleNetV1 to 0.05) and decay it by 0.1 at 100 and 150 epochs. On Tiny-ImageNet and ImageNet datasets, we train models for 100 epochs. Learning rate is initialized to 0.1 and decayed every 30 epochs.

\subsection{Comparasions with State-of-the-art Methods}
\textbf{Results on CIFAR-100.} We evaluate our method on three similar and two different structures of teacher-student pairs. We show the results on Table~\ref{tab:cifar-100}. From this table, we have the following observations. 1) The methods that exploit the logit-based knowledge (\ie, KD, ESKD, TAKD) show superiority over those feature-based methods. Such a phenomenon has also been found in CRD~\cite{tian2019contrastive}. We claim that logit-based knowledge has not yet been fully exploited, which is overlooked in recent
state-of-the-art feature-based methods. 2) Our \method is well generalized to different teacher-student pairs. On average, our \method achieves higher accuracy than the vanilla student by 3.62\%, ranging from 2.86\% to 4.31\%. 3) Our \method performs better than all of the compared methods. Specifically, our method evenly outperforms vanilla KD and two related methods (\ie, ESKD and TAKD) by 1.58\%, 1.71\%, 0.93\% on all teacher-student pairs. Note that ESKD is inferior to vanilla KD in many teacher-student combinations. The main reason is that they lost a lot of teacher's knowledge. On the contrary, we improve the student's capacity to effectively exploit the teacher's knowledge.

\textbf{Results on CINIC-10.} In this experiment, we evaluate our method on two teacher-student pairs. As shown in Table~\ref{tab:cinic-10}, our method achieves the best results on all pairs. Specifically, our \method outperforms vanilla student by 1.91\% on average, which is a relatively large improvement on CINIC-10 dataset. These results demonstrate the effectiveness of our method on more complex dataset.

\textbf{Results on Tiny-ImageNet.} In this experiment, we evaluate \method with the pair of ResNet50 and ResNet18. From Table~\ref{tab:tiny-imagenet}, we have the following observations. First, most distillation methods outperform the vanilla student by a large margin. It is worth noting that the students in many distillation methods even achieve higher accuracy than the teacher model. Such a phenomenon shows the importance and effectiveness of knowledge distillation on a more complex and challenging dataset. Second, our \method performs better than all of the compared methods by at least 1.02\% in terms of Top-1 accuracy. These results show the effectiveness and generalization of our \method on more challenging datasets.

\textbf{Results on ImageNet.}
In this experiment, we compare our \method with state-of-the-art methods by using the pair of ResNet34 and ResNet18. Note that TAKD~\cite{mirzadeh2019improved} does not provide comparable ImageNet results. From Figure~\ref{tab:ImageNet}, our proposed \method achieves highest accuracy. Specifically, our \method reduces the performance gap between the teacher and the student from 3.56\% to 1.98\%, a 44\% relative improvement. These results demonstrate the scalability of our proposed \method.

\begin{figure}[t]
    \centering
	\includegraphics[width=1\linewidth]{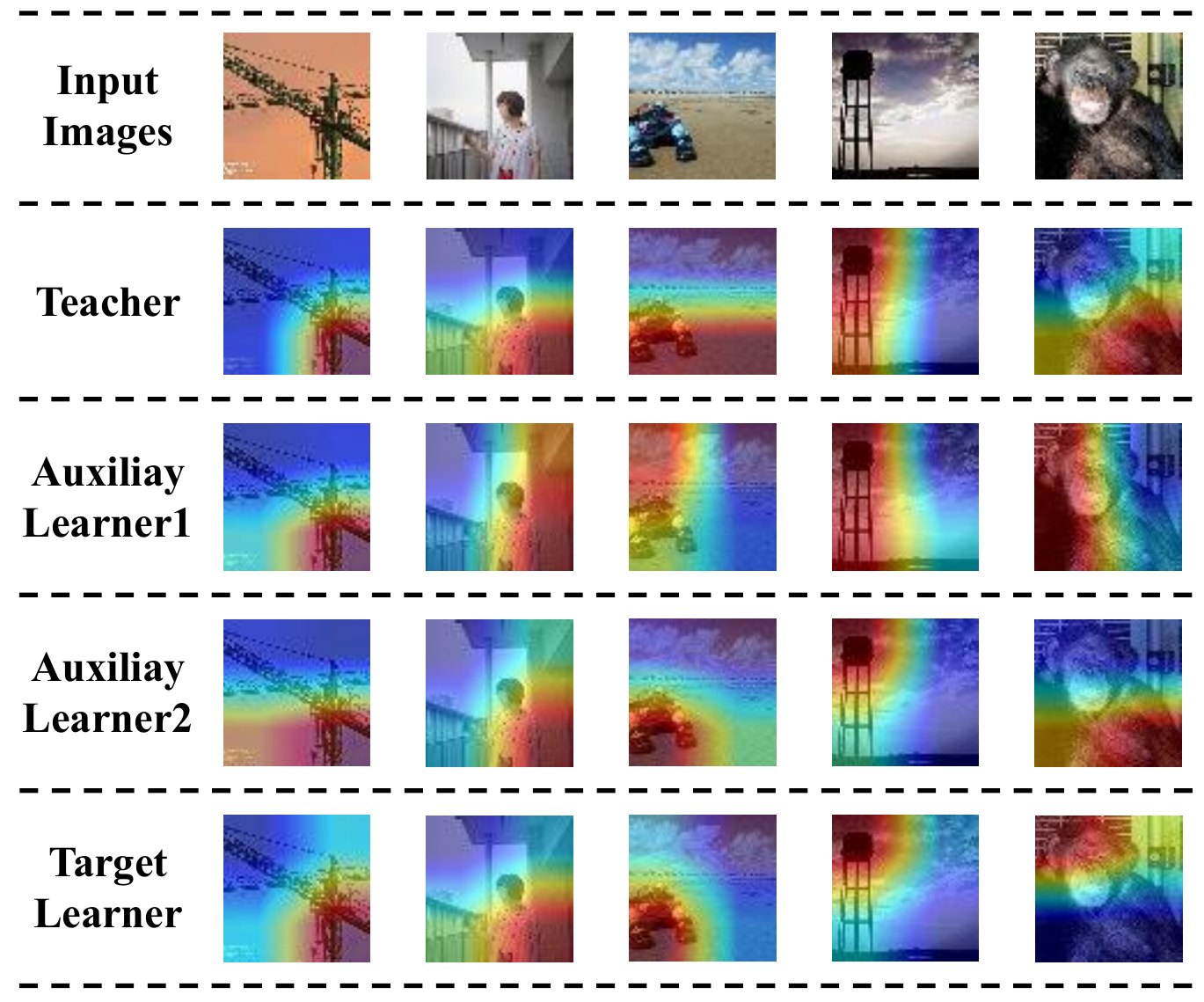}
	\caption{Gram-CAM visualization on Tiny-ImageNet. }
	\label{fig:cam}
\end{figure}

\subsection{Effect of Adversarial Collaborative Module}
\textbf{Qualitative Analysis.} In this experiment, we analyze how the adversarial collaborative module works by Grad-CAM visualization. We randomly select several samples from the testing set on Tiny-ImageNet. We show the qualitative results of different parts (\ie, teacher, target learner and auxiliary learners) of \method framework in Figure~\ref{fig:cam}. We observe that 1) The regions highlighted by different auxiliary learners are somewhat different. Interestingly, there always exists one learner activates similar region to the teacher (\eg, auxiliary learner1 in the $1st$ and $4th$ column). 2) The target learner captures more semantic-related information than auxiliary learners even the teacher model. For example, in the $3rd$ column, the target learner successfully puts its attention on the whole target objective while the teacher distracts part of its attention to the background. The main reason is that the auxiliary learners capture diverse information, which helps the target learner learn more accurate and robust class-discriminative representation.

\textbf{Effect of proposed modules.} In this experiment, we conduct experiments to investigate the effectiveness of proposed modules. We show the results in Table~\ref{tab:ablation-loss}. 1) The performance significantly boost when we apply the auxiliary learners to collaboratively learn the knowledge from the teacher and then transfer to the student (\ie, collaboration). 2) Equipping with the attention or adversarial module alone improves the performance, which means that the attention and the adversarial module effectively strength the collaboration and diversity, respectively. 3) When equipped with all proposed modules, the student achieves highest accuracy. These results demonstrate the effectiveness of our proposed adversarial collaborative module.

\begin{table}[t]
\begin{center}
\resizebox{0.45\textwidth}{!}{
\begin{tabular}{ccc|c}
            \toprule
            Collaboration & Attention module & Adversarial module & Accuracy (\%) \\
            \midrule
             &  &  & 70.73 \\
            $\surd$ &  &  & 71.67\\
            $\surd$ & $\surd$ &  & 72.02 \\
            $\surd$ &  & $\surd$ & 71.99 \\
            $\surd$& $\surd$ & $\surd$ & \textbf{72.38} \\
			\bottomrule
		\end{tabular}}
\end{center}
	\caption{Effect of different modules.}
		\label{tab:ablation-loss}
\end{table}
% We choose resnet110 and resnet20 as teacher and student, respectively. We report the Top-1 accuracy ($\%$) for resnet20 on CIFAR-100.

\begin{table}[t]
\begin{center}
\begin{tabular}{cc|c}
            \toprule
            learner1 & learner2 & Accuracy (\%) \\
            \midrule
            D3B1 & D3B1 & 72.02 \\ 
            D3B1 & D2B2 & \textbf{72.38} \\
            D2B2 & D2B2 & 71.64 \\
            D2B2 & D1B3 & 72.16 \\
            D1B3 & D1B3 & 71.40 \\
			\bottomrule
		\end{tabular}
\end{center}
	\caption{Effect of different learners. "D$N$B$M$`` denotes that we add an auxiliary learner with $M$ increased feature extractor blocks to the $Nth$ down-sampling layer.}
		\label{tab:branch-position}
\end{table}	

\textbf{Effect of different learners.} 
% In this paper, we attach auxiliary learners to collaboratively learn the teacher's knowledge. Thus, how to design the learners is important. 
In this experiment, we investigate the effect of different learners. We use "D$N$B$M$`` to denote that we add a learner with $M$ increased feature extractor blocks to the $Nth$ down-sampling layer. From Table~\ref{tab:branch-position}, we have the following observations. 1) Different learners all lead to performance improvement compared with vanilla KD (70.73\%), which shows the effectiveness and robustness of collaborative module. 2) Attaching learners to different down-sampling layers shows superiority 
over same layers. The main reason is that they effectively utilize different semantic information of the student, improving the diversity of learners. 3) ``D3B1 with D2B2" achieves higher accuracy than ``D2B2 with D1B3". The main reason is that the student shares more low-level layers with learners, leading to a more stable training process.

\section{Conclusion}
In this paper, we have proposed an Adversarial Collaborative Knowledge Distillation (\method) method that effectively improves the performance of the student model in the knowledge distillation task. To this end, we have constructed a student model with multiple auxiliary learners. Based on the reinforced student model, we have proposed an adversarial collaborative learning strategy. Specifically, we have proposed an attention module to promote collaboration between auxiliary learners and used discriminators to distinguish the output distribution of auxiliary learners to enhance representational diversity. Extensive experiments on four benchmarks have shown that our \method obtains state-of-the-art performance.

{\small
\bibliographystyle{ieee_fullname}
\bibliography{egbib}

\begin{thebibliography}{10}\itemsep=-1pt

\bibitem{ahn2019variational}
Sungsoo Ahn, Shell~Xu Hu, Andreas Damianou, Neil~D Lawrence, and Zhenwen Dai.
\newblock Variational information distillation for knowledge transfer.
\newblock In {\em IEEE Conf. Comput. Vis. Pattern Recog.}, pages 9163--9171,
  2019.

\bibitem{chen2020online}
Defang Chen, Jian-Ping Mei, Can Wang, Yan Feng, and Chun Chen.
\newblock Online knowledge distillation with diverse peers.
\newblock In {\em AAAI}, pages 3430--3437, 2020.

\bibitem{chen2020cross}
Defang Chen, Jian-Ping Mei, Yuan Zhang, Can Wang, Zhe Wang, Yan Feng, and Chun
  Chen.
\newblock Cross-layer distillation with semantic calibration.
\newblock {\em AAAI}, 2021.

\bibitem{chin2020towards}
Ting-Wu Chin, Ruizhou Ding, Cha Zhang, and Diana Marculescu.
\newblock Towards efficient model compression via learned global ranking.
\newblock In {\em IEEE Conf. Comput. Vis. Pattern Recog.}, pages 1518--1528,
  2020.

\bibitem{cho2019efficacy}
Jang~Hyun Cho and Bharath Hariharan.
\newblock On the efficacy of knowledge distillation.
\newblock In {\em IEEE Conf. Comput. Vis. Pattern Recog.}, pages 4794--4802,
  2019.

\bibitem{chung2020feature}
Inseop Chung, SeongUk Park, Jangho Kim, and Nojun Kwak.
\newblock Feature-map-level online adversarial knowledge distillation.
\newblock In {\em International Conference on Machine Learning}, pages
  2006--2015. PMLR, 2020.

\bibitem{dietterich2000ensemble}
Thomas~G Dietterich.
\newblock Ensemble methods in machine learning.
\newblock In {\em International workshop on multiple classifier systems}, pages
  1--15. Springer, 2000.

\bibitem{dillenbourg1999collaborative}
Pierre Dillenbourg.
\newblock {\em Collaborative learning: Cognitive and computational approaches.
  advances in learning and instruction series.}
\newblock ERIC, 1999.

\bibitem{goodfellow2014generative}
Ian~J Goodfellow, Jean Pouget-Abadie, Mehdi Mirza, Bing Xu, David Warde-Farley,
  Sherjil Ozair, Aaron Courville, and Yoshua Bengio.
\newblock Generative adversarial networks.
\newblock {\em Adv. Neural Inform. Process. Syst.}, 2014.

\bibitem{Goodfellow2014GenerativeAN}
Ian~J. Goodfellow, Jean Pouget-Abadie, Mehdi Mirza, Bing Xu, David
  Warde-Farley, Sherjil Ozair, Aaron~C. Courville, and Yoshua Bengio.
\newblock Generative adversarial networks.
\newblock {\em Adv. Neural Inform. Process. Syst.}, abs/1406.2661, 2014.

\bibitem{DBLP:conf/cvpr/GuoWWYLHL20}
Qiushan Guo, Xinjiang Wang, Yichao Wu, Zhipeng Yu, Ding Liang, Xiaolin Hu, and
  Ping Luo.
\newblock Online knowledge distillation via collaborative learning.
\newblock In {\em IEEE Conf. Comput. Vis. Pattern Recog.}, pages 11017--11026,
  2020.

\bibitem{han2015deep}
Song Han, Huizi Mao, and William~J Dally.
\newblock Deep compression: Compressing deep neural networks with pruning,
  trained quantization and huffman coding.
\newblock {\em Int. Conf. Learn. Represent.}, 2016.

\bibitem{he2016deep}
Kaiming He, Xiangyu Zhang, Shaoqing Ren, and Jian Sun.
\newblock Deep residual learning for image recognition.
\newblock In {\em IEEE Conf. Comput. Vis. Pattern Recog.}, pages 770--778,
  2016.

\bibitem{he2017channel}
Yihui He, Xiangyu Zhang, and Jian Sun.
\newblock Channel pruning for accelerating very deep neural networks.
\newblock In {\em Int. Conf. Comput. Vis.}, pages 1389--1397, 2017.

\bibitem{hinton2015distilling}
Geoffrey Hinton, Oriol Vinyals, and Jeff Dean.
\newblock Distilling the knowledge in a neural network.
\newblock {\em arXiv preprint arXiv:1503.02531}, 2015.

\bibitem{hu2018squeeze}
Jie Hu, Li Shen, and Gang Sun.
\newblock Squeeze-and-excitation networks.
\newblock In {\em IEEE Conf. Comput. Vis. Pattern Recog.}, pages 7132--7141,
  2018.

\bibitem{huang2017densely}
Gao Huang, Zhuang Liu, Laurens Van Der~Maaten, and Kilian~Q Weinberger.
\newblock Densely connected convolutional networks.
\newblock In {\em IEEE Conf. Comput. Vis. Pattern Recog.}, pages 4700--4708,
  2017.

\bibitem{jacob2018quantization}
Benoit Jacob, Skirmantas Kligys, Bo Chen, Menglong Zhu, Matthew Tang, Andrew
  Howard, Hartwig Adam, and Dmitry Kalenichenko.
\newblock Quantization and training of neural networks for efficient
  integer-arithmetic-only inference.
\newblock In {\em IEEE Conf. Comput. Vis. Pattern Recog.}, pages 2704--2713,
  2018.

\bibitem{jin2019knowledge}
Xiao Jin, Baoyun Peng, Yichao Wu, Yu Liu, Jiaheng Liu, Ding Liang, Junjie Yan,
  and Xiaolin Hu.
\newblock Knowledge distillation via route constrained optimization.
\newblock In {\em Int. Conf. Comput. Vis.}, pages 1345--1354, 2019.

\bibitem{kornblith2019similarity}
Simon Kornblith, Mohammad Norouzi, Honglak Lee, and Geoffrey Hinton.
\newblock Similarity of neural network representations revisited.
\newblock {\em International Conference on Machine Learning}, 2019.

\bibitem{lee2015deeply}
Chen-Yu Lee, Saining Xie, Patrick Gallagher, Zhengyou Zhang, and Zhuowen Tu.
\newblock Deeply-supervised nets.
\newblock In {\em Artificial intelligence and statistics}, pages 562--570.
  PMLR, 2015.

\bibitem{li2020online}
Zheng Li, Ying Huang, Defang Chen, Tianren Luo, Ning Cai, and Zhigeng Pan.
\newblock Online knowledge distillation via multi-branch diversity enhancement.
\newblock In {\em ACCV}, 2020.

\bibitem{lin2017focal}
Tsung-Yi Lin, Priya Goyal, Ross Girshick, Kaiming He, and Piotr Doll{\'a}r.
\newblock Focal loss for dense object detection.
\newblock In {\em Int. Conf. Comput. Vis.}, pages 2980--2988, 2017.

\bibitem{liu2016ssd}
Wei Liu, Dragomir Anguelov, Dumitru Erhan, Christian Szegedy, Scott Reed,
  Cheng-Yang Fu, and Alexander~C Berg.
\newblock Ssd: Single shot multibox detector.
\newblock In {\em Eur. Conf. Comput. Vis.}, pages 21--37. Springer, 2016.

\bibitem{Liu2021CTSF}
Yanxia Liu, Anni Chen, Hongyu Shi, Sijuan Huang, Wanjia Zheng, Zhiqiang Liu,
  Qin Zhang, and Xin Yang.
\newblock Ct synthesis from mri using multi-cycle gan for head-and-neck
  radiation therapy.
\newblock {\em Computerized medical imaging and graphics : the official journal
  of the Computerized Medical Imaging Society}, 91:101953, 2021.

\bibitem{liu2021semi}
Zhiqiang Liu, Yanxia Liu, and Chengkai Huang.
\newblock Semi-online knowledge distillation.
\newblock {\em Brit. Mach. Vis. Conf.}, 2021.

\bibitem{ma2018shufflenet}
Ningning Ma, Xiangyu Zhang, Hai-Tao Zheng, and Jian Sun.
\newblock Shufflenet v2: Practical guidelines for efficient cnn architecture
  design.
\newblock In {\em Eur. Conf. Comput. Vis.}, pages 116--131, 2018.

\bibitem{mirzadeh2019improved}
Seyed-Iman Mirzadeh, Mehrdad Farajtabar, Ang Li, Nir Levine, Akihiro Matsukawa,
  and Hassan Ghasemzadeh.
\newblock Improved knowledge distillation via teacher assistant.
\newblock {\em AAAI}, 2020.

\bibitem{peng2019correlation}
Baoyun Peng, Xiao Jin, Jiaheng Liu, Dongsheng Li, Yichao Wu, Yu Liu, Shunfeng
  Zhou, and Zhaoning Zhang.
\newblock Correlation congruence for knowledge distillation.
\newblock In {\em Int. Conf. Comput. Vis.}, pages 5007--5016, 2019.

\bibitem{rastegari2016xnor}
Mohammad Rastegari, Vicente Ordonez, Joseph Redmon, and Ali Farhadi.
\newblock Xnor-net: Imagenet classification using binary convolutional neural
  networks.
\newblock In {\em Eur. Conf. Comput. Vis.}, pages 525--542. Springer, 2016.

\bibitem{redmon2016you}
Joseph Redmon, Santosh Divvala, Ross Girshick, and Ali Farhadi.
\newblock You only look once: Unified, real-time object detection.
\newblock In {\em IEEE Conf. Comput. Vis. Pattern Recog.}, pages 779--788,
  2016.

\bibitem{ren2015faster}
Shaoqing Ren, Kaiming He, Ross Girshick, and Jian Sun.
\newblock Faster r-cnn: Towards real-time object detection with region proposal
  networks.
\newblock In {\em Adv. Neural Inform. Process. Syst.}, pages 91--99, 2015.

\bibitem{romero2014fitnets}
Adriana Romero, Nicolas Ballas, Samira~Ebrahimi Kahou, Antoine Chassang, Carlo
  Gatta, and Yoshua Bengio.
\newblock Fitnets: Hints for thin deep nets.
\newblock {\em Int. Conf. Learn. Represent.}, 2015.

\bibitem{sandler2018mobilenetv2}
Mark Sandler, Andrew Howard, Menglong Zhu, Andrey Zhmoginov, and Liang-Chieh
  Chen.
\newblock Mobilenetv2: Inverted residuals and linear bottlenecks.
\newblock In {\em IEEE Conf. Comput. Vis. Pattern Recog.}, pages 4510--4520,
  2018.

\bibitem{shen2019meal}
Zhiqiang Shen, Zhankui He, and Xiangyang Xue.
\newblock Meal: Multi-model ensemble via adversarial learning.
\newblock In {\em AAAI}, volume~33, pages 4886--4893, 2019.

\bibitem{song2018collaborative}
Guocong Song and Wei Chai.
\newblock Collaborative learning for deep neural networks.
\newblock {\em Adv. Neural Inform. Process. Syst.}, 31:1832--1841, 2018.

\bibitem{szegedy2015going}
Christian Szegedy, Wei Liu, Yangqing Jia, Pierre Sermanet, Scott Reed, Dragomir
  Anguelov, Dumitru Erhan, Vincent Vanhoucke, and Andrew Rabinovich.
\newblock Going deeper with convolutions.
\newblock In {\em IEEE Conf. Comput. Vis. Pattern Recog.}, pages 1--9, 2015.

\bibitem{tian2019contrastive}
Yonglong Tian, Dilip Krishnan, and Phillip Isola.
\newblock Contrastive representation distillation.
\newblock {\em Int. Conf. Learn. Represent.}, 2020.

\bibitem{tung2019similarity}
Frederick Tung and Greg Mori.
\newblock Similarity-preserving knowledge distillation.
\newblock In {\em Int. Conf. Comput. Vis.}, pages 1365--1374, 2019.

\bibitem{wu2016quantized}
Jiaxiang Wu, Cong Leng, Yuhang Wang, Qinghao Hu, and Jian Cheng.
\newblock Quantized convolutional neural networks for mobile devices.
\newblock In {\em IEEE Conf. Comput. Vis. Pattern Recog.}, pages 4820--4828,
  2016.

\bibitem{xie2015holistically}
Saining Xie and Zhuowen Tu.
\newblock Holistically-nested edge detection.
\newblock In {\em Int. Conf. Comput. Vis.}, pages 1395--1403, 2015.

\bibitem{Xie2020DeepTQ}
Zheng Xie, Zhiquan Wen, Jing Liu, Zhiqiang Liu, Xixian Wu, and Mingkui Tan.
\newblock Deep transferring quantization.
\newblock In {\em Eur. Conf. Comput. Vis.}, 2020.

\bibitem{yao2020knowledge}
Anbang Yao and Dawei Sun.
\newblock Knowledge transfer via dense cross-layer mutual-distillation.
\newblock In {\em Eur. Conf. Comput. Vis.}, pages 294--311. Springer, 2020.

\bibitem{ye2019student}
Jingwen Ye, Yixin Ji, Xinchao Wang, Kairi Ou, Dapeng Tao, and Mingli Song.
\newblock Student becoming the master: Knowledge amalgamation for joint scene
  parsing, depth estimation, and more.
\newblock In {\em IEEE Conf. Comput. Vis. Pattern Recog.}, pages 2829--2838,
  2019.

\bibitem{you2018learning}
Shan You, Chang Xu, Chao Xu, and Dacheng Tao.
\newblock Learning with single-teacher multi-student.
\newblock In {\em AAAI}, volume~32, 2018.

\bibitem{yuan2020revisiting}
Li Yuan, Francis~EH Tay, Guilin Li, Tao Wang, and Jiashi Feng.
\newblock Revisiting knowledge distillation via label smoothing regularization.
\newblock In {\em IEEE Conf. Comput. Vis. Pattern Recog.}, pages 3903--3911,
  2020.

\bibitem{zagoruyko2016wide}
Sergey Zagoruyko and Nikos Komodakis.
\newblock Wide residual networks.
\newblock {\em Brit. Mach. Vis. Conf.}, 2016.

\bibitem{Zagoruyko2017PayingMA}
Sergey Zagoruyko and Nikos Komodakis.
\newblock Paying more attention to attention: Improving the performance of
  convolutional neural networks via attention transfer.
\newblock {\em Int. Conf. Learn. Represent.}, abs/1612.03928, 2017.

\bibitem{zhang2020task}
Linfeng Zhang, Yukang Shi, Zuoqiang Shi, Kaisheng Ma, and Chenglong Bao.
\newblock Task-oriented feature distillation.
\newblock {\em Adv. Neural Inform. Process. Syst.}, 33, 2020.

\bibitem{zhang2020amln}
Xiaobing Zhang, Shijian Lu, Haigang Gong, Zhipeng Luo, and Ming Liu.
\newblock Amln: adversarial-based mutual learning network for online knowledge
  distillation.
\newblock In {\em Eur. Conf. Comput. Vis.}, pages 158--173. Springer, 2020.

\bibitem{zhang2018shufflenet}
Xiangyu Zhang, Xinyu Zhou, Mengxiao Lin, and Jian Sun.
\newblock Shufflenet: An extremely efficient convolutional neural network for
  mobile devices.
\newblock In {\em IEEE Conf. Comput. Vis. Pattern Recog.}, pages 6848--6856,
  2018.

\bibitem{Zhang2018DeepML}
Y. Zhang, T. Xiang, Timothy~M. Hospedales, and H. Lu.
\newblock Deep mutual learning.
\newblock {\em IEEE Conf. Comput. Vis. Pattern Recog.}, pages 4320--4328, 2018.

\bibitem{zhao2017pyramid}
Hengshuang Zhao, Jianping Shi, Xiaojuan Qi, Xiaogang Wang, and Jiaya Jia.
\newblock Pyramid scene parsing network.
\newblock In {\em IEEE Conf. Comput. Vis. Pattern Recog.}, pages 2881--2890,
  2017.

\bibitem{Lan2018KnowledgeDB}
Xiatian Zhu, Shaogang Gong, et~al.
\newblock Knowledge distillation by on-the-fly native ensemble.
\newblock In {\em Adv. Neural Inform. Process. Syst.}, pages 7517--7527, 2018.

\bibitem{zhuang2018discrimination}
Zhuangwei Zhuang, Mingkui Tan, Bohan Zhuang, Jing Liu, Yong Guo, Qingyao Wu,
  Junzhou Huang, and Jinhui Zhu.
\newblock Discrimination-aware channel pruning for deep neural networks.
\newblock In {\em Adv. Neural Inform. Process. Syst.}, pages 875--886, 2018.

\end{thebibliography}
}

\end{document}